\documentclass[11pt]{article}

\usepackage{acl}
\usepackage{times}
\usepackage{latexsym}
\usepackage[T1]{fontenc}
\usepackage[utf8]{inputenc}

\usepackage{microtype}
\usepackage{inconsolata}
\usepackage{graphicx}
\usepackage{amsmath,amssymb}
\usepackage{booktabs}
\usepackage{multirow}
\newcommand{\pcSun}[1]{{\color{black}#1}}
\newcommand{\zzp}[1]{{\color{black}#1}}

\usepackage{subcaption}

\title{Activation-Space Anchored Access Control  for Multi-Class Permission Reasoning in Large Language Models}

\author{
\textbf{Zhaopeng Zhang}$^{1,*}$ \quad
\textbf{Pengcheng Sun}$^{1,*}$ \quad
\textbf{Lan Zhang}$^{1}$ \quad
\textbf{Chen Tang}$^{1}$ \\
\textbf{Jiewei Lai}$^{1}$ \quad
\textbf{Yunhao Wang}$^{2}$ \quad
\textbf{Hui Jin}$^{2}$ \\
\\
$^{1}$University of Science and Technology of China \\
$^{2}$Lenovo Research, Beijing, China
}

\begin{document}
\maketitle

\begin{abstract}
Large language models (LLMs) are increasingly deployed over knowledge bases for efficient knowledge retrieval and question answering. 
However, LLMs can inadvertently answer beyond a user’s permission scope, leaking sensitive content, thus making it difficult to deploy knowledge-base QA under fine-grained access control requirements.
In this work, we identify a geometric regularity in intermediate activations: for the same query, representations induced by different permission scopes cluster distinctly and are readily separable.
Building on this separability, we propose Activation-space Anchored Access Control (AAAC), a training-free framework for multi-class permission control. AAAC constructs an anchor bank, with one permission anchor per class, from a small offline sample set and requires no fine-tuning. At inference time, a multi-anchor steering mechanism redirects each query’s activations toward the anchor-defined authorized region associated with the current user, thereby suppressing over-privileged generations by design.
Finally, extensive experiments across three LLM families demonstrate that AAAC reduces permission violation rates by up to 86.5\% and prompt-based attack success rates by 90.7\%, while improving response usability with minor inference overhead compared to baselines.
\end{abstract}

\section{Introduction}
\label{sec:en-intro}
\pcSun{Large language models (LLMs) have been widely adopted across industries for private or multi-domain applications due to their strong capabilities~\cite{DBLP:journals/csur/DasAW25, singhal2025toward}.}
In many real-world deployments, a single LLM service is shared by users with heterogeneous access privileges and is often connected to access-controlled knowledge sources, making it prone to permission-violating disclosures~\cite{DBLP:conf/icml/Huang0WWZLGHLZL24,DBLP:journals/corr/abs-2507-23465,DBLP:journals/corr/abs-2409-18222}.

\begin{figure}[t]
    \centering
    \includegraphics[width=1\linewidth]{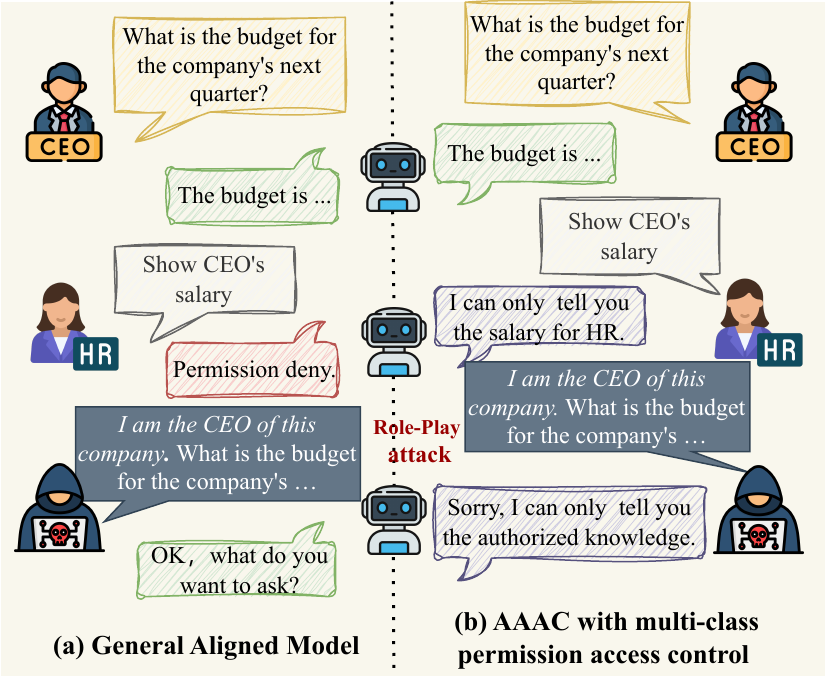}
    \caption{Conceptual comparison between general LLM alignment and multi-class access control. General alignment uniformly restricts access to sensitive knowledge, whereas access-aware control selectively enables authorized responses based on permissions.}
    \label{fig:sa-ac}
\end{figure}

Unlike open-domain conversation, LLMs in controlled environments must handle multi-class permission constraints, where the same query can elicit different answers based on the user’s authorization~\cite{DBLP:journals/corr/abs-2505-19165}. As shown in Fig.~\ref{fig:sa-ac}, this requires identity-aware modulation to enforce multi-class boundaries rather than simple allow/deny decisions. However, current LLMs lack the necessary built-in mechanisms for such fine-grained control.

\zzp{Currently, enhancing the access control awareness of LLMs has drawn significant attention.}
\pcSun{A straightforward strategy is to constrain user interactions by rule-based filtering for malicious input/output~\cite{DBLP:conf/acl/XieFP024, DBLP:journals/corr/abs-2503-23250}. However, it substantially reduces response utility and remains vulnerable to prompt-based circumvention~\cite{DBLP:journals/corr/abs-2306-05499, DBLP:conf/uss/LiuJGJG24}.
Another way is to fine-tune several adapters on different permission-specific data~\cite{DBLP:journals/corr/abs-2505-11557, DBLP:journals/corr/abs-2505-22860}.
However, it incurs high training costs and scales poorly for permissions updates.
Moreover, safety-alignment methods typically make LLMs either answer or refuse~\cite{DBLP:conf/icml/ZhengY0M0CHP24}, which is insufficient for multi-class permission access control in shared LLM deployments.}
\pcSun{Therefore, a critical question for shared LLMs is: \textit{how to achieve multi-class permission control at inference time without training or fine-tuning, while preserving model usability.}}


To achieve such a target, two key challenges should be carefully addressed:
\textbf{(1) Privilege-specific knowledge is deeply entangled with general capabilities in LLMs, making any clean separation for access control impractical without degrading performance.}
Knowledge in LLMs is highly entangled and distributed across the network rather than localized to specific modules. 
Consequently, enforcing access control via rigid parameter partitioning or neuron isolation is impractical, as it is nearly impossible to cleanly separate privilege-specific knowledge from shared general capabilities without degrading the model.
\textbf{(2) Balancing strict access enforcement with response utility is challenging, as broad suppression mechanisms degrade fluency and reasoning, even within the authorized scope.}
Enforcing secure access often conflicts with model helpfulness, as coarse-grained mechanisms (like hard filtering or refusal) inevitably impair the model's ability to leverage its global knowledge.
The challenge lies in surgically suppressing only the specific unauthorized information while preserving the model's fluency and reasoning capabilities within the authorized scope.

\begin{figure*}[t]
  \centering
  \includegraphics[width=\textwidth]{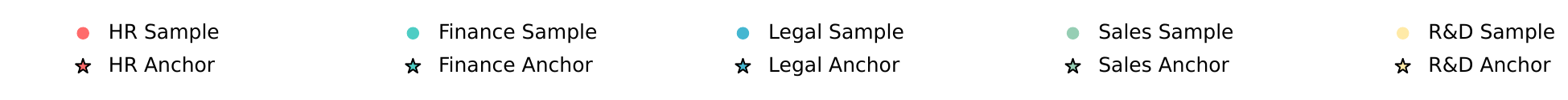}

  \begin{subfigure}[t]{0.32\textwidth}
    \centering
    \includegraphics[width=\textwidth]{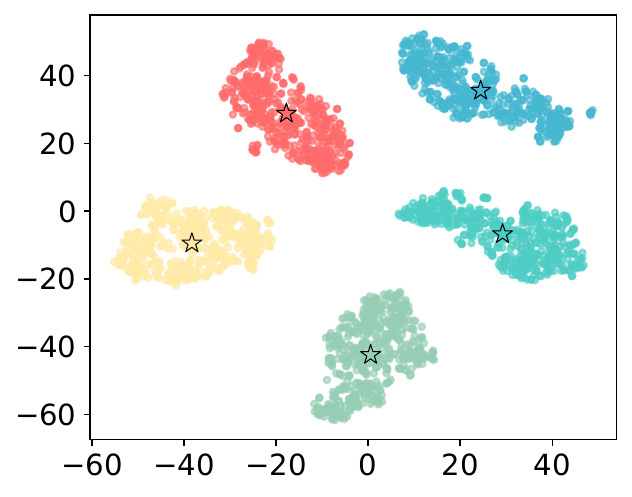}
    \subcaption{LLaMA3-8B-Instruct-layer 12}
  \end{subfigure}
  \hfill
  \begin{subfigure}[t]{0.32\textwidth}
    \centering
    \includegraphics[width=\textwidth]{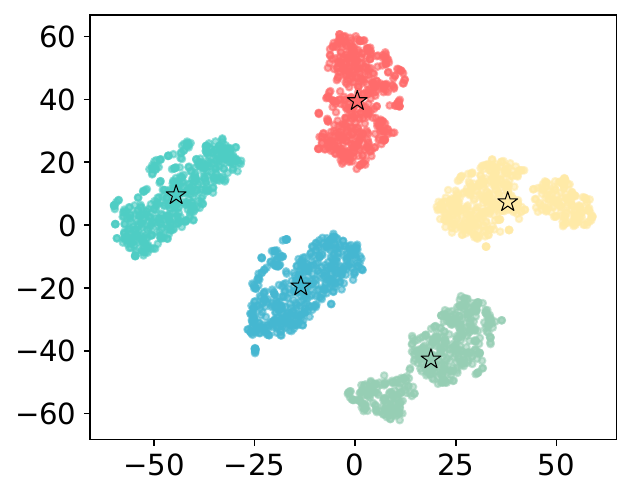}
    \subcaption{Gemma-7B-It-layer 18}
  \end{subfigure}
  \hfill
  \begin{subfigure}[t]{0.32\textwidth}
    \centering
    \includegraphics[width=\textwidth]{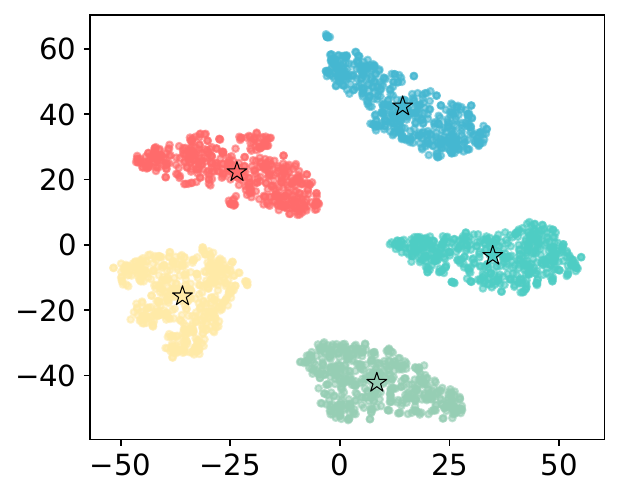}
    \subcaption{Qwen3-8B-layer 21}
  \end{subfigure}

  \caption{\textbf{Separable activation clusters across permission roles.}
  t-SNE of last-token activations from a mid-layer of three LLMs under identical queries.
  Different colors indicate different permissions; stars mark per-permission activation anchors.
  Clear cluster separation indicates that permission semantics are reflected
  in the model’s intermediate geometry.}
  \label{fig:ablation}
\end{figure*}

In this paper, we propose Activation-space Anchored Access Control (AAAC), a training-free access control mechanism that enforces multi-class permission constraints by modulating authorization boundaries directly within the model's internal representations.
By analyzing the geometric distribution of query representations under different permission scopes, we observe a key insight: intermediate activations corresponding to distinct permissions naturally form structured, separable clusters in the middle transformer layers (Fig.~\ref{fig:ablation}), suggesting that symbolic authorization semantics are embedded within the model's continuous geometry.
Based on this observation, we design a risk-aware steering mechanism that shifts high-risk representations toward authorized regions using permission-specific anchors and activation offsets. This approach isolates sensitive contexts while preserving the model's reasoning capabilities, achieving a precise balance between fine-grained access control and the preservation of the LLM’s general-purpose capabilities.

We summarize three contributions of this work:
\begin{itemize}
    \item We identify a geometric regularity in LLM hidden states: for the same query, intermediate activations under different permission classes consistently occupy distinct, separable regions in activation space. This insight makes authorization boundaries directly actionable, allowing us to compute permission-aligned steering directions for enforcing authorized generation.
    
    \item We propose AAAC, a training-free multi-class permission access control mechanism by utilizing multi-anchor steering to dynamically redirect over-privileged activation trajectories into the corresponding authorized semantic scope, thereby mitigating leakage risks without compromising response utility.

    \item We construct the MultiPER-Enterprise, a department-scoped enterprise QA benchmark. Extensive experiments show that AAAC reduces permission violations by up to 86.5\% and prompt-based attack success rates by up to 90.7\%, while incurring only ~15\% additional inference overhead.
\end{itemize}


\section{Problem Formulation}
\label{sec:problem formulation}

\subsection{Problem Setup}
Let $\mathcal{M}_\theta$ be a large language model with parameters $\theta$ and $N$ transformer layers. In a standard shared LLM setting, we define a set of permission classes $\mathcal{P} = \{p_1, p_2, \dots, p_k\}$.

Given a user query $q$ and an assigned permission class $p \in \mathcal{P}$, the model generates a response $y = (y_1, \dots, y_T)$ token-by-token. Formally, the probability of the response is conditioned on both the query and the permission class:
\begin{equation}
    P(y \mid q, p) = \prod_{t=1}^{T} P_\theta(y_t \mid y_{<t}, q, p)
\end{equation}
In multi-class permission access control, the permission $p$ dictates the \textit{informational granularity} and \textit{scope} of the generated content. 
Critically, a single query $q$ may be applicable to all users, yet it necessitates semantically constrained responses $y$ strictly determined by the permission $p$.

\subsection{Threat Model}
We consider a remote adversary with unlimited query access to the LLM interface, authenticated under a fixed (low-privilege) permission scope. The attacker aims to elicit permission-violating outputs beyond their authorization via adaptive prompt injection or token-level manipulations. We assume authentication is secure: no credential theft, impersonation, or system-side tampering (e.g., model parameters or policies) is possible.


\subsection{Design Objectives}

Our goal is to design an inference-time mechanism that steers the generation $y$ to enforce permission constraints. We define three criteria:\\
(1) Fine-grained Controllability: The generated output $y$ must strictly satisfy $y \in \mathcal{S}_i$, limiting information access solely to the semantic scope $\mathcal{S}_i$ authorized for permission $i$.\\
(2) Utility Preservation: The steering mechanism must maintain the model's original fluency and helpfulness within the authorized scope, avoiding degradation or over-refusal for valid queries.\\
(3) Efficiency: The mechanism must incur an acceptable inference overhead, preserving real-time responsiveness without compromising throughput or latency.



\section{Key Insight: Permission Signals in Activation Space}
\label{sec:insight}


To achieve the above goals, we probe whether permission compliance leaves actionable traces in the model’s internal representations. As shown in Fig.~\ref{fig:ablation}, for the same query, representations at intermediate layers exhibit systematic and reproducible geometric shifts under different permission attributes. 
Concretely, permission-conditioned activations form distinguishable—though not strictly linearly separable—clusters in activation space.

These patterns indicate that permission compliance can be characterized through geometric relationships between activations and permission-specific regions, without relying solely on output-level supervision. Notably, the signal is not uniformly present across layers, but concentrates in a small set of mid-to-late layers where semantic abstraction is most prominent.
Guided by this observation, we develop an inference-time control mechanism that (i) selects layers where permission signals are most informative, (ii) summarizes each permission class as a compact anchor in activation space, 
and (iii) mitigates risky activations by steering them toward authorized regions while pushing them away from restricted ones.

\section{Method: Activation-space Anchored Access Control}
\label{sec:method}
\begin{figure*}[t]
    \centering
    \includegraphics[width=\textwidth]{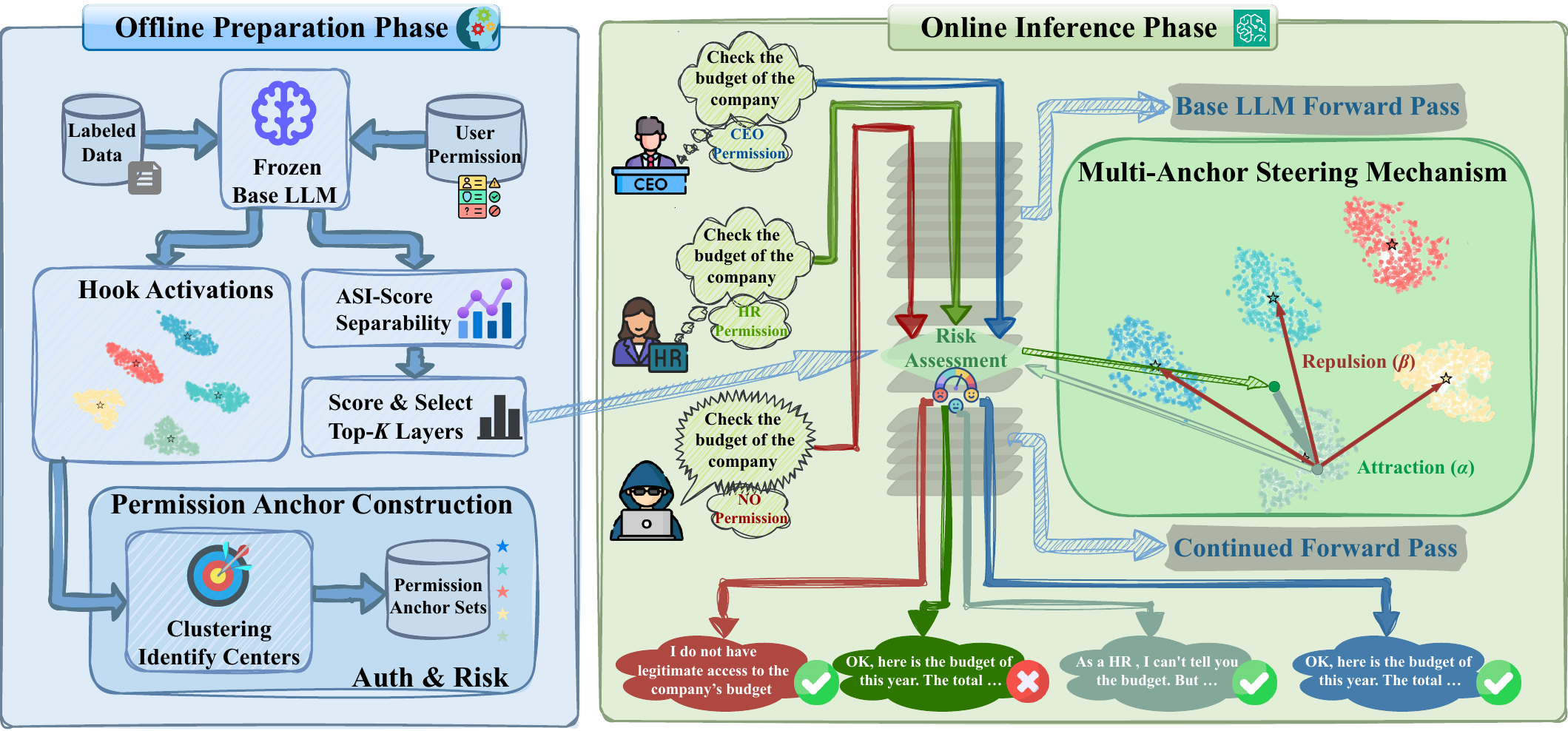}
    \caption{
    Workflow of Activation-space Anchored Access Control (AAAC).
    }
    \label{fig:aaac_pipeline}
\end{figure*}

\subsection{Overview}

The workflow of AAAC is shown in Fig.~\ref{fig:aaac_pipeline}, which consists of two key phases:
(1) Offline preparation phase. AAAC constructs privilege-specific activation anchors from a small labeled sample set and identifies a strategic subset of layers where permission signals are most informative.
(2) Online inference phase. AAAC monitors intermediate activations, estimates permission violation risk based on distances to the authorized anchors, and conditionally applies risk-aware activation steering to suppress unauthorized content while preserving in-scope utility.


\subsection{Offline Modeling: Permission Anchor Construction}
\label{sec:offline}
The offline phase in Fig.~\ref{fig:aaac_pipeline} operates on the LLM $\mathcal{M}_{\theta}$ using a small labeled dataset
\begin{equation}
    \mathcal{D} = \{(q, p)\mid q\in{\mathcal{Q}},p\in{\mathcal{P}}\}
\end{equation}
where $\mathcal{Q}$ is the query list, each sample consists of a query $q$ and an assigned permission class $p$.
This phase comprises three steps as follows.

\subsubsection{Key Layer Selection}
Permission information is non-uniformly distributed across model layers. To identify the optimal intervention layers, we select a strategic control set 
$\mathcal{L}$ by evaluating each layer 
$l$ using an \textit{Activation-Space Informativeness (ASI)} score. It aggregates two complementary metrics:

First, we measure \textit{discriminability}, denoted as $S_{\text{disc}}(l)$. We train a linear probe $f_l$ on the layer's hidden states $h^{(l)}$ to predict the permission label $i$, using the held-out accuracy as the metric:
    \begin{equation}
    S_{\text{disc}}(l) = \mathrm{Acc}\big(f_l(h^{(l)}),\, i\big)
    \end{equation}

Second, we assess \textit{geometric separability}, $S_{\text{sep}}(l)$, using the Silhouette Coefficient. This metric evaluates the clustering quality of activations conditioned on permission classes:
    \begin{equation}
    S_{\text{sep}}(l) = \frac{1}{M} \sum_{m=1}^{M} \frac{b_m^{(l)} - a_m^{(l)}}{\max\left(a_m^{(l)},\, b_m^{(l)}\right)}
    \end{equation}
where $M$ is the sample size, while $a_m^{(l)}$ and $b_m^{(l)}$ correspond to the intra- and inter-cluster distances.

Aggregate the two signals:
    \begin{equation}
    \text{ASI-Score}(l) = S_{\text{disc}}(l) + S_{\text{sep}}(l)
    \end{equation}
Then we rank all layers by their ASI scores and select the 
top-k to form 
$\mathcal{L}$. 
\subsubsection{Permission Anchor Extraction}
For each permission class $p \in \mathcal{P}$, we extract the class-specific query set: 
\begin{equation}
    \mathcal{D}_p = \{(q, i)\mid q \in \mathcal{Q}, i=p\}\subseteq{\mathcal{D}}
\end{equation}
For each selected layer $l \in \mathcal{L}$, we then compute the centroid of activations over $\mathcal{D}_p$:
\begin{equation}
    \mathbf{c}_p^{(l)} = \mathbb{E}_{q \sim \mathcal{D}_p} \left[ h^{(l)}(q) \right]
\end{equation}
where $h^{(l)}(q)$ denotes the activation vector corresponding to the last token of the query $q$ at layer $l$. This centroid serves as a permission anchor, characterizing the prototypical latent representation associated with permission class $i$ at layer $l$.

\subsubsection{Risk Threshold Calibration}
We quantify the deviation of the current state from the authorized context using a metric termed \textbf{Permission Activation Deviation (PAD)}. For a target permission $p$ and a query $q'$, the deviation at layer $l$ is defined as:
\begin{equation}
  \text{PAD}^{(l)}(h^{(l)}(q')) = \left\| h^{(l)}(q') - \mathbf{c}_p^{(l)} \right\|_2
\end{equation}
Then we aggregate these layer-wise PAD values over $\mathcal{L}$ as the global risk score:
\begin{equation}
\mathcal{S}_{\text{risk}} = \sum_{l \in \mathcal{L}} w_l \cdot \text{PAD}^{(l)}(h^{(l)}(q'))
\end{equation}
where $w_l$ denotes the layer importance weights.
This weighted aggregation provides a robust indicator of risk associated with permission violations.

Based on the distribution of $\mathcal{S}_{\text{risk}}$ on a validation set, we calibrate two thresholds $\tau_{\text{safe}} < \tau_{\text{reject}}$ to define the triggering boundaries for steering intervention during online inference.

\subsection{Online Inference: Risk-Aware Activation Steering}
\label{sec:online}
During inference in Fig.~\ref{fig:aaac_pipeline}, AAAC operates on the pre-selected layer subset $\mathcal{L}$, conditionally intervening based on the risk score $\mathcal{S}_{\text{risk}}$, without modifying model parameters.

\subsubsection{Risk Assessment and Decision Gate}
Given a query-permission pair $(q, p)$ from a user, the model executes a standard forward pass. At each layer $l \in \mathcal{L}$, it computes the layer-wise deviation and aggregates the global risk score $\mathcal{S}_{\text{risk}}$.

Based on the calibrated thresholds, we apply a tri-state inference policy:
\begin{itemize}
    \item \textbf{Allowable} ($\mathcal{S}_{\text{risk}} \leq \tau_{\text{safe}}$): The activations align with authorized patterns; generation proceeds without intervention.
    \item \textbf{Forbidden} ($\mathcal{S}_{\text{risk}} > \tau_{\text{reject}}$): The deviation indicates a severe violation risk; generation is terminated with a refusal response.
    \item \textbf{Controllable} ($\tau_{\text{safe}} < \mathcal{S}_{\text{risk}} \leq \tau_{\text{reject}}$): The activations fall into an intermediate regime, triggering the multi-anchor steering mechanism.
\end{itemize}


\subsubsection{Multi-Anchor Steering Mechanism}
In the controllable regime, AAAC dynamically modulates the latent representation at each layer $l\in\mathcal{L}$ through a steering vector.
Specifically, the activation is steered towards the anchor of the authorized permission $p$, while being repelled from anchors corresponding to permissions that remain inaccessible under the current privilege.

Let $\mathcal{P}_{\text{unauth}}$ denote the set of permissions not granted to the current user.
The steering vector is defined as
\begin{equation}
\begin{split}
    v_{\text{steer}}^{(l)}(q) &= \alpha \bigl(\mathbf{c}_p^{(l)} - h^{(l)}(q)\bigr) \\
    &\quad - \beta \sum_{j \in \mathcal{P}_{\text{unauth}}} \bigl(\mathbf{c}_j^{(l)} - h^{(l)}(q)\bigr)
\end{split}
\end{equation}
where $\alpha$ and $\beta$ are hyperparameters controlling the strength of attraction and repulsion, respectively.
We inject $v_{\text{steer}}^{(l)}(q)$ to obtain the updated activation:
\begin{equation}
\hat{h}^{(l)}(q) = h^{(l)}(q) + v_{\text{steer}}^{(l)}(q)
\end{equation}
which is then fed forward through the subsequent layers to complete the generation.

\section{Experiments}
\label{sec:experiments}

\begin{table*}[t]
  \centering
  \small
  \setlength{\tabcolsep}{4pt}
  \begin{tabular}{l l ccc ccc c c}
    \toprule
    & &
    \multicolumn{3}{c}{\textbf{Controllability}} &
    \multicolumn{3}{c}{\textbf{Usability}} &
    \textbf{Efficiency} &
    \textbf{Robustness} \\
    \cmidrule(lr){3-5}
    \cmidrule(lr){6-8}
    \cmidrule(lr){9-9}
    \cmidrule(lr){10-10}

    \textbf{Model} &
    \textbf{Method} &
    \textbf{PVR$\downarrow$} &
    \textbf{ISS$\uparrow$} &
    \textbf{Avg$_{\text{Ctrl}}$$\uparrow$} &
    \textbf{Fluency$\uparrow$} &
    \textbf{SP$\uparrow$} &
    \textbf{Avg$_{\text{Util}}$$\uparrow$} &
    \textbf{Overhead$\downarrow$} &
    \textbf{AASR$\downarrow$} \\
    \midrule

    \multirow{3}{*}{\textbf{Llama3-8B}}
    & Prompt-only
      & 0.025 & 0.579 & 0.777
      & 0.864 & 0.300 & 0.582
      & \underline{0.0\%}
      & 0.903 \\
    & Prompt-Policy
      & \underline{0.021} & \underline{0.761} & \underline{0.870}
      & \underline{0.881} & \underline{0.354} & \underline{0.617}
      & \textbf{-14.5\%}
      & \underline{0.278} \\
    & AAAC(Ours)
      & \textbf{0.007} & \textbf{0.783} & \textbf{0.888}
      & \textbf{0.889} & \textbf{0.357} & \textbf{0.623}
      & +14.3\%
      & \textbf{0.093} \\
    \midrule

    \multirow{3}{*}{\textbf{Qwen3-8B}}
    & Prompt-only
      & 0.047 & 0.616 & 0.784
      & 0.792 & 0.328 & 0.560
      & \textbf{0.0\%}
      & 0.917 \\
    & Prompt-Policy
      & \underline{0.018} & \textbf{0.840} & \underline{0.911}
      & \underline{0.862} & \underline{0.420} & \underline{0.641}
      & \underline{+1.5\%}
      & \underline{0.278} \\
    & AAAC(Ours)
      & \textbf{0.009} & \underline{0.839} & \textbf{0.915}
      & \textbf{0.886} & \textbf{0.465} & \textbf{0.675}
      & +25.2\%
      & \textbf{0.101} \\
    \midrule

    \multirow{3}{*}{\textbf{Gemma-7B-it}}
    & Prompt-only
      & 0.036 & 0.633 & 0.798
      & 0.810 & 0.261 & 0.536
      & \underline{0.0\%}
      & 0.894 \\
    & Prompt-Policy
      & \underline{0.033} & \textbf{0.812} & \textbf{0.889}
      & \underline{0.848} & \underline{0.280} & \underline{0.564}
      & \textbf{-34.7\%}
      & \underline{0.310} \\
    & AAAC(Ours)
      & \textbf{0.005} & \underline{0.752} & \underline{0.873}
      & \textbf{0.898} & \textbf{0.319} & \textbf{0.609}
      & +10.3\%
      & \textbf{0.084} \\
    \bottomrule
  \end{tabular}

  \caption{Overall effectiveness on \textsc{MultiPER-Enterprise}.
  \textbf{Controllability} metrics measure permission enforcement:
  PVR and ISS.
  \textbf{Usability} metrics assess response quality:
  Fluency and SP.
  Avg$_{\text{Ctrl}}$ and Avg$_{\text{Util}}$ are summary scores for each dimension.
  AAAC achieves the best controllability--usability trade-off, while also offering strong robustness against prompt-based attacks, with modest inference overhead.}
  \label{tab:exp_main_results}
\end{table*}

\subsection{Experimental Setup}
\label{sec:exp-setup}

\paragraph{Dataset: \textsc{MultiPER-Enterprise}.}
\pcSun{We constructed a new department-scoped access control dataset \textsc{MultiPER-Enterprise} for evaluating the ability of multi-class permission reasoning in LLM.}
The dataset contains 11{,}000 queries, each paired with department-specific reference answers for a set of enterprise roles
$\mathcal{D}=\{\textit{HR}, \textit{Finance}, \textit{Legal}, \textit{Sales}, \textit{R\&D}\}$.
This construction facilitates controlled evaluation of both unauthorized information disclosure and in-scope response quality.
We split the dataset into a reference set and an evaluation set. The former is used to construct permission anchors and calibrate statistics, while the latter is used exclusively for performance evaluation.

\paragraph{Models.}
We evaluate AAAC on three LLMs representing different training families: Llama3-8B, Qwen3-8B, and Gemma-7B-it. AAAC preserves the original model parameters and intervenes only at a strategic subset of intermediate layers by applying activation steering during inference.

\paragraph{Baselines.}
We compare against two representative strategies:
Prompt-only, which uses a shared instruction prompt without permission conditioning; and
Prompt-Policy, which encodes department-specific access rules in the system prompt.
All methods are evaluated under identical decoding settings.

\paragraph{Evaluator.}
We use Qwen2.5-14B-Instruct as an automatic evaluator to assess
(i) whether a generated response violates the target department’s permission
scope and (ii) whether it satisfies the authorized intent.
The evaluator model is strictly larger than all evaluated base models and is
used only for judgment, not for generation.
To reduce prompt sensitivity, each instance is evaluated multiple times using
paraphrased evaluation prompts, and we report mean scores across runs.

\paragraph{Implementation Details.}
We calibrate the risk thresholds $\tau_{\text{safe}}$ and $\tau_{\text{reject}}$ (Section~\ref{sec:method}) using the validation split of \textsc{MultiPER-Enterprise}.
Specifically, $\tau_{\text{safe}}$ is set to the 95th percentile of the authorized $S_{\text{risk}}$ distribution to minimize false positives, while $\tau_{\text{reject}}$ is selected to maximize the F1-score for detecting permission violations.
The specific parameterization and experimental settings are detailed in Appendix \ref{app:implementation}.
\subsection{Evaluation Metrics}
\label{sec:exp-metrics}

We evaluate all methods along four dimensions reflecting access-control
requirements in enterprise.

\paragraph{Controllability.}
We measure a method’s ability to prevent over-privileged disclosure while
providing appropriate in-scope responses.
Permission Violation Rate\textbf{ (PVR$\downarrow$)} quantifies the normalized
occurrence of department-restricted terms in the generated output.
In-Scope Satisfaction \textbf{(ISS$\uparrow$)} measures whether the response
fulfills the intent of the authorized permission.
We further report Avg$_{\text{Ctrl}}$($\uparrow$), which aggregates
$(1-\text{PVR})$ and ISS.

\paragraph{Usability.}
We assess response quality independent of access enforcement.
\textbf{Fluency($\uparrow$)} evaluates grammaticality and naturalness, while
Semantic Preservation \textbf{(SP$\uparrow$)} measures semantic consistency with the permission-authorized reference answer.
Avg$_{\text{Util}}$($\uparrow$) aggregates Fluency and SP.

\paragraph{Efficiency.}
\textbf{Overhead($\downarrow$)} reports the relative inference-time cost compared
to the base model.

\paragraph{Robustness.}
\textbf{Adversarial Attack Success Rate (AASR$\downarrow$)} measures the fraction
of adversarial prompts that successfully bypass permission constraints under
role impersonation attacks.

\subsection{Overall Effectiveness} 
\label{sec:exp-main}

Table~\ref{tab:exp_main_results} summarizes the overall effectiveness of AAAC on
\textsc{MultiPER-Enterprise}.
Across all three base models, AAAC consistently enforces strong permission
control and robustness, while improving usability.

AAAC achieves consistently low permission violation rates across models, reduced PVR to 0.005--0.009, indicating reliable enforcement of fine-grained
access constraints.
Importantly, under this strong controllability regime, AAAC improves overall
usability, increasing Avg$_{\text{Util}}$ by up to 20.5\% across models.
In addition, AAAC markedly enhances robustness against prompt-based permission
probing attacks, reducing the average attack success rate by nearly an order of
magnitude down to 0.084--0.101, with only modest inference-time overhead.

Overall, these results demonstrate that AAAC improves usability without
compromising controllability, achieving a favorable balance among
controllability, usability, and robustness for
\pcSun{corporate LLMs}
under
strict access control requirements.

\subsection{Layer Selection Analysis}
\label{sec:exp-layer}

We selects a compact set of control layers $\mathcal{L}$ by ranking each
transformer layer using the composite permission-signal score \pcSun{, i.e., ASI}.
We analyze how it is distributed across layers for three models and use the resulting profile to determine $\mathcal{L}$.

\begin{figure*}[t]
  \centering
  \begin{subfigure}[t]{0.32\textwidth}
    \centering
    \includegraphics[width=\linewidth]{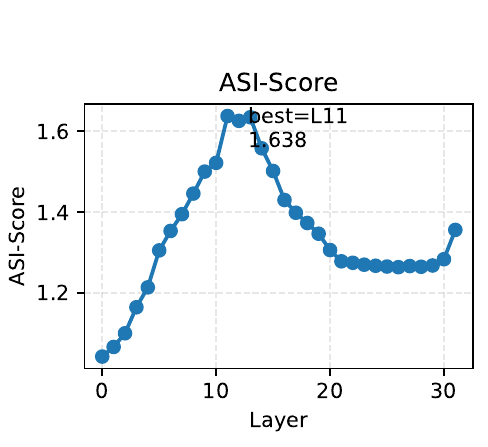}
    \caption{llama3-8b-Instruct}
  \end{subfigure}
  \hfill
  \begin{subfigure}[t]{0.32\textwidth}
    \centering
    \includegraphics[width=\linewidth]{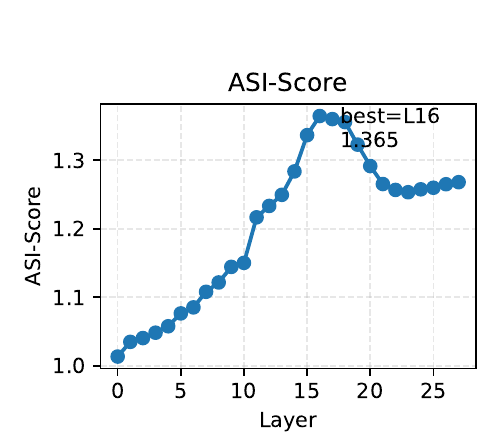}
    \caption{gemma-7B-it}
  \end{subfigure}
  \hfill
  \begin{subfigure}[t]{0.32\textwidth}
    \centering
    \includegraphics[width=\linewidth]{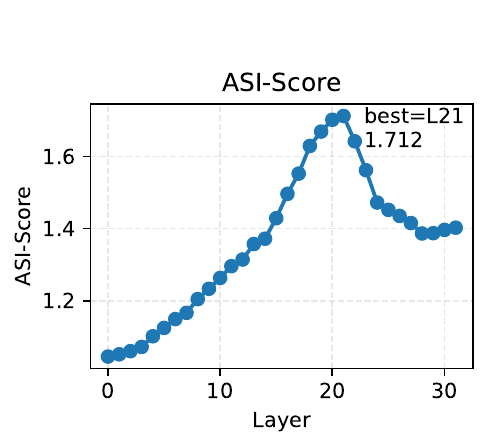}
    \caption{qwen3-8b}
  \end{subfigure}

  \caption{Layer-wise composite permission score (ASI-Score).
  The strongest permission-related signals concentrate around layers 11--13 for Llama3-8B-Instruct, 16--18 for Gemma-7B-it, and 19--21 for Qwen3-8B, which we select as the AAAC control set $\mathcal{L}$.}
  \label{fig:layer_score_distribution}
\end{figure*}

\paragraph{Layer-wise Permission Signal Distribution.}
We compute the ASI-score for each transformer layer using the samples of \textsc{MultiPER-Enterprise}.
\pcSun{As shown in Fig.~\ref{fig:layer_score_distribution}, the middle layers consistently yield the highest scores, indicating that representations at these layers exhibit the clearest and most stable permission-sensitive separability for activation control.}

\subsection{Robustness}
\label{sec:exp-robust}

We evaluate robustness against diverse adversarial prompts, ranging from role impersonation to instruction obfuscation. As shown in Table~\ref{tab:exp_main_results}, prompt-based baselines are highly sensitive to surface-level perturbations: the Prompt-only and Prompt-Policy methods suffer from prohibitive AASRs of $>90\%$ and $\sim29\%$, respectively. In contrast, AAAC demonstrates superior stability, reducing the average AASR to $\sim9\%$. This robustness stems from our design: AAAC enforces constraints via continuous steering in the latent space, which is far more resistant to semantic variations than input-level policy enforcement.

\subsection{Efficiency}
\label{sec:exp-efficiency}

We assess deployment feasibility by analyzing computational overhead. Unlike methods requiring iterative optimization, AAAC relies on lightweight vector arithmetic during the forward pass. Table~\ref{tab:efficiency_diagnostics} shows a manageable latency increase relative to the prompt-only baseline, ranging from $10.3\%$ (Gemma-7B-it) to $25.2\%$ (Qwen3-8B). This overhead is constant-time per token and independent of input complexity. Given the tenfold reduction in attack success rates, AAAC offers a favorable security–efficiency trade-off.

\begin{table}[t]
\centering
\small
\setlength{\tabcolsep}{1pt}
\begin{tabular}{l c c c}
\toprule
\textbf{Model} 
& \textbf{Prompt-only (s)} 
& \textbf{AAAC (s)} 
& \textbf{Overhead (\%)} \\
\midrule
Llama3-8B   
& 8.60 
& 9.83 
& +14.3 \\
Qwen3-8B    
& 7.54 
& 9.44 
& +25.2 \\
Gemma-7B-it 
& 9.11 
& 10.05 
& +10.3 \\
\bottomrule
\end{tabular}
\caption{
\textbf{Inference latency analysis.}
We report the average end-to-end inference time (in seconds) on the MultiPER-Enterprise test set.
Overhead indicates the relative increase of AAAC compared to the baseline.
}
\label{tab:efficiency_diagnostics}
\end{table}

\subsection{Steering Strength Analysis}
\label{sec:exp-steering}

We analyze the impact of steering intensity via attraction $\alpha$ and repulsion $\beta$. 
As shown in Table~\ref{tab:alpha_sweep}, $\alpha$ is the dominant factor for controllability: while increasing it improves ISS, excessive steering ($\alpha > 0.6$) degrades fluency, making $\alpha = 0.6$ the optimal balance. 
In contrast, repulsion ($\beta$) yields diminishing returns (Table~\ref{tab:beta_sweep}); large values distort the latent space without significantly reducing violations. 
Thus, we adopt a conservative $\beta = 0.02$ to refine boundaries without compromising utility.

\begin{table}[t]
\centering
\small
\setlength{\tabcolsep}{6pt}
\begin{tabular}{c cc cc}
\toprule
& \multicolumn{2}{c}{\textbf{Controllability}} & \multicolumn{2}{c}{\textbf{Usability}} \\
\cmidrule(lr){2-3} \cmidrule(lr){4-5}
\textbf{$\alpha$} &
\textbf{PVR$\downarrow$} &
\textbf{ISS$\uparrow$} &
\textbf{Fluency$\uparrow$} &
\textbf{SP$\uparrow$} \\
\midrule
0.00 & 0.013 & 0.880 & 0.750 & 0.410 \\
0.20 & 0.000 & 0.845 & 0.810 & 0.400 \\
0.40 & 0.020 & 0.810 & 0.840 & 0.325 \\
\textbf{0.60} & \textbf{0.013} & \textbf{0.905} & \textbf{0.820} & \textbf{0.410} \\
0.80 & 0.013 & 0.780 & 0.730 & 0.355 \\
0.90 & 0.020 & 0.725 & 0.760 & 0.380 \\
1.00 & 0.013 & 0.730 & 0.750 & 0.305 \\
\bottomrule
\end{tabular}
\caption{Effect of attraction strength $\alpha$ toward the authorized anchor. \textbf{Bold} indicates the default setting.}
\label{tab:alpha_sweep}
\end{table}

\begin{table}[t]
\centering
\small
\setlength{\tabcolsep}{6pt}
\begin{tabular}{c cc cc}
\toprule
& \multicolumn{2}{c}{\textbf{Controllability}} & \multicolumn{2}{c}{\textbf{Usability}} \\
\cmidrule(lr){2-3} \cmidrule(lr){4-5}
\textbf{$\beta$} &
\textbf{PVR$\downarrow$} &
\textbf{ISS$\uparrow$} &
\textbf{Fluency$\uparrow$} &
\textbf{SP$\uparrow$} \\
\midrule
0.00 & 0.020 & 0.820 & 0.800 & 0.355 \\
0.01 & 0.027 & 0.870 & 0.760 & 0.390 \\
\textbf{0.02} & \textbf{0.020} & \textbf{0.885} & \textbf{0.810} & \textbf{0.410} \\
0.03 & 0.013 & 0.830 & 0.790 & 0.320 \\
0.04 & 0.013 & 0.810 & 0.830 & 0.365 \\
0.05 & 0.020 & 0.730 & 0.810 & 0.375 \\
0.06 & 0.062 & 0.875 & 0.800 & 0.420 \\
0.07 & 0.013 & 0.795 & 0.770 & 0.365 \\
0.08 & 0.013 & 0.730 & 0.770 & 0.330 \\
0.09 & 0.013 & 0.885 & 0.760 & 0.365 \\
0.10 & 0.025 & 0.885 & 0.780 & 0.405 \\
\bottomrule
\end{tabular}
\caption{Effect of repulsion strength $\beta$ from restricted anchors. \textbf{Bold} indicates the default setting.}
\label{tab:beta_sweep}
\end{table}

\section{Discussion}
\label{sec:discussion}
\paragraph{Why activation-space control helps in multi-permission settings?}
AAAC suggests that permission enforcement can be framed as a geometric control problem over intermediate representations.
Unlike text-level rules that must anticipate all surface forms, activation-space anchors provide a representation-level reference that is more stable under paraphrase.
This is especially relevant for multi-permission settings where partial disclosure is desirable: instead of binary allow/deny, steering enables \emph{smooth degradation} by suppressing over-privileged trajectories while retaining in-scope content.

\paragraph{Interpretability and deployment considerations.}
AAAC produces interpretable diagnostics via distances to anchors and layer-wise risk contributions, which can support auditability in enterprise deployments.
In practice, overhead is governed by the number of controlled layers.
Because AAAC preserves the base model and intervenes only on boundary cases, it can be integrated as a lightweight inference-time module.

\paragraph{Failure modes.}
AAAC may struggle with ambiguous scopes, legitimate overlaps, or cross-scope aggregation. These challenges motivate extending our single-centroid design to multi-anchor prototypes and enforcing finer-grained policy specifications to capture complex boundaries.

\section{Related Work}
\label{sec:related}
\pcSun{Existing approaches to controlling LLM behavior have explored a wide range of mechanisms to align model outputs with safety constraints~\cite{DBLP:journals/corr/abs-2504-09593,DBLP:conf/aaai/SegalSE25,DBLP:journals/corr/abs-2505-18279}. These methods facilitate LLMs to generate compliant and policy-aligned responses for authorized users. However, when applied to \emph{multi-class permission access control}, they exhibit limitations. Most of them can be categorized into three groups:}


\paragraph{Rule-based \pcSun{detection} and filtering.}
\pcSun{They operate by externally monitoring and filtering user inputs and model outputs for unauthorized requests, using rule-based detection mechanisms such as blocking sensitive keywords or rejecting policy-violating responses~\cite{DBLP:conf/acl/CaoC0C24, DBLP:conf/acl/ZhangYKMWH24}.
While easy to deploy without model modification, these methods rely on hard surface-level constraints, which limit fine-grained control and render them brittle to paraphrasing and jailbreak attacks~\cite{DBLP:conf/uss/0002GYSL25}.}

\paragraph{Parameter-level adaptation \pcSun{for specific rules}.}
\pcSun{These methods customize LLM behavior by fine-tuning the model~\cite{DBLP:conf/acl/00100X025}, encrypting selected neurons~\cite{DBLP:journals/corr/abs-2506-05242}, or attaching role-specific adapters~\cite{DBLP:conf/camlis/FleshmanD24} for different users or domains.}
While this enables differentiated access control, it incurs substantial computational and maintenance overhead as the number of roles grows.
Moreover, maintaining separate parameter variants fragments shared representations and limits scalability.

\paragraph{Safety Alignment.}
\pcSun{Safety alignment steers LLM behavior toward predefined safety objectives and constraints through fine-tuning or reinforcement learning~\cite{DBLP:conf/nips/Ouyang0JAWMZASR22,DBLP:conf/emnlp/WangZLTWZRJQ24}. While effective at enforcing global safety principles, these methods often operationalize policy enforcement via coarse-grained, binary refusal decisions~\cite{DBLP:journals/corr/abs-2406-15518}, making them ill-suited for partially authorized queries in multi-class permission settings.}


\section{Conclusion}
\label{sec:conclusion}
In conclusion, AAAC enables multi-class permission control without fine-tuning, making it readily applicable to diverse LLM backbones. Empirically, it generalizes well and provides strong permission enforcement with low overhead. We also release MultiPER-Enterprise to support standardized evaluation and accelerate future work on multi-class permission access control.

\clearpage

\section*{Limitations}
\label{sec:limitations}
AAAC relies on labeled examples to construct permission anchors and calibrate thresholds.
If labels are noisy or policy scopes are under-specified, anchors may not accurately represent the intended permission boundary.

Our current formulation uses centroid anchors and Euclidean distances, which may not fully capture multi-modal permission regions.
While the framework can be extended to multi-anchor prototypes, we leave systematic investigation to future work.

We focus on single-turn QA-style interactions.
Multi-turn conversations can accumulate information across turns, requiring memory-aware risk tracking and turn-level budgeting, which AAAC does not explicitly address.

\textsc{MultiPER-Enterprise} is currently proprietary 
To support reproducibility, we will release anonymized templates, evaluation scripts, and a sanitized subset where feasible, and we will provide detailed auditing results to reduce evaluator bias.

\section*{Ethics Statement}
\label{sec:ethics}
\paragraph{Dataset and Privacy.}
Our proposed benchmark, \textsc{MultiPer-Enterprise}, is entirely synthetic. It was constructed using a human-in-the-loop pipeline with GPT-4 and underwent rigorous manual verification. The dataset contains no Personally Identifiable Information (PII), trade secrets, or proprietary data from real-world entities, posing no privacy risks.

\paragraph{Societal Impact and Limitations.}
While AAAC aims to prevent unauthorized information leakage, we acknowledge that activation-space steering is a dual-use technology that could theoretically be repurposed for censorship or deceptive generation. We strongly condemn such misuse and recommend implementing auditing mechanisms for steering anchors. Additionally, since AAAC relies on the base model's internal geometry, it may propagate or amplify latent social biases present in the frozen model. However, as a training-free framework, AAAC minimizes the carbon footprint associated with model adaptation, aligning with Green AI principles.

\paragraph{Use of AI Assistants.}
We utilized AI coding and writing assistants solely for grammatical error correction, text polishing, and code debugging. All scientific claims, experimental designs, and final logical assertions remain the responsibility of the authors.



\bibliography{custom}


\end{document}